%% file: main.tex
\documentclass[10pt,twocolumn,letterpaper]{article}

\usepackage{cvpr}              

\input{preamble}

\definecolor{cvprblue}{rgb}{0.21,0.49,0.74}
\usepackage[pagebackref,breaklinks,colorlinks,citecolor=cvprblue]{hyperref}
\usepackage{multirow}
\usepackage{microtype}
\usepackage{graphicx}
\usepackage{booktabs} 
\usepackage{amsmath}
\usepackage{amssymb}
\usepackage{mathtools}
\usepackage{amsthm}
\theoremstyle{plain}

\theoremstyle{definition}

\theoremstyle{remark}

\usepackage{algorithm}  
\usepackage{algorithmic}  

\newcommand{\hr}[1]{\textcolor{black}{#1}}

\newcommand{\tgreen}[1]{\textcolor{green}{#1}}
\newcommand{\up}[1]{\tgreen{$\uparrow$#1}}

\newcommand{\hrup}[1]{\hr{$\uparrow$#1}} 
\newcommand{\hrdown}[1]{\hr{$\downarrow$#1}}

\def\paperID{ECV-30} 
\def\confName{CVPR}
\def\confYear{2024}

\title{ShiftAddAug: Augment Multiplication-Free Tiny Neural Network with Hybrid Computation}

\author{Yipin Guo, Zihao Li, Yilin Lang, Qinyuan Ren\\
College of Control Science and Engineering, Zhejiang University\\
{\tt\small \{guoyipin, lzh\_jeong, langyilin, renqinyuan\}@zju.edu.cn}
}
\begin{document}
\maketitle
\input{sec/0_abstract}\footnote{Accepted by 2024 CVPR Workshop : Efficient Deep Learning for Computer Vision} 
\vspace{-2em}
\input{sec/1_intro}
\input{sec/2_rela}
\input{sec/3_method}

\input{sec/4_exp}

\input{sec/5_conclusion}
{
    \small
    \bibliographystyle{ieeenat_fullname}
    \bibliography{reference}
}

\newpage
\appendix
\onecolumn



\end{document}

%% file: preamble.tex
\usepackage[dvipsnames]{xcolor}


%% file: sec/0_abstract.tex
\begin{abstract}
Operators devoid of multiplication, such as Shift and Add, have gained prominence for their compatibility with hardware.  
However, neural networks (NNs) employing these operators typically exhibit lower accuracy compared to conventional NNs with identical structures. \textbf{ShiftAddAug} uses costly multiplication to augment efficient but less powerful multiplication-free operators, improving performance without any inference overhead. It puts a ShiftAdd tiny NN into a large multiplicative model and encourages it to be trained as a sub-model to obtain additional supervision. In order to solve the weight discrepancy problem between hybrid operators, a new weight sharing method is proposed.
Additionally, a novel two stage neural architecture search is used to obtain better augmentation effects for smaller but stronger multiplication-free tiny neural networks.
The superiority of ShiftAddAug is validated through experiments in image classification and semantic segmentation, consistently delivering noteworthy enhancements. Remarkably,  it secures up to a 4.95\% increase in accuracy on the CIFAR100 compared to its directly trained counterparts, even surpassing the performance of multiplicative NNs. 
\end{abstract}

%% file: sec/1_intro.tex
\section{Introduction}
\label{sec:intro}


\indent 
\indent The application of deep neural networks (DNNs) on resource-constrained platforms is still limited due to their huge energy requirements and computational costs. To obtain a small model deployed on edge devices, the commonly used techniques are pruning\cite{pruning1,pruning2}, quantization\cite{quantization1,quantization2}, and knowledge distillation\cite{knowledge_distillation}. However, the NNs designed by the above works are all based on multiplication. The common hardware design practice in digital signal processing tells that multiplication can be replaced by bit-wise shifts and additions\cite{xue1986adaptive,Gwee2009shiftadd_multiplier} to achieve faster speed and lower energy consumption. Introducing this idea into NNs design, DeepShift\cite{elhoushi2021deepshift} and AdderNet\cite{chen2021addernet} proposed ShiftConv operator and AddConv operator respectively.

This paper takes one step further along the direction of multiplication-free neural networks, proposing a method to augment tiny multiplication-free NNs by hybrid computation, which significantly improves accuracy without any inference overhead. Considering that multiplication-free operators cannot restore all the information from the original operator, tiny NNs employing ShiftAdd calculations exhibit pronounced under-fitting. Drawing inspiration from NetAug\cite{NetAug}, ShiftAddAug chooses to build a larger hybrid computing NN for training and sets the multiplication-free part as the target model used in inference and deployment. The stronger multiplicative part as augmentation is expected to push the target multiplication-free model to a better condition.

In augmented training, the hybrid operators share weights, but because different operators have varying weight distributions, effective weights for multiplication may not suit shift or add operations. This led us to develop a strategy for heterogeneous weight sharing in augmentation.

Furthermore, since NetAug limits augmentation to width, ShiftAddAug aims to extend this by exploring depth and operator change levels. Thus, we adopt a two-step neural architecture search strategy to find highly efficient, multiplication-free tiny neural networks.

ShiftAddAug is evaluated on MCU-level tiny models. Compared to the multiplicative NNs, directly trained multiplication-free NNs can obtain considerable speed improvement (\hr{2.94$\times$ to 3.09$\times$}) and energy saving (\hrdown{67.75\%$\sim$69.09\%}) at the cost of reduced accuracy. ShiftAddAug consistently enhances accuracy(\hrup{1.08\%$\sim$4.95\%}) while maintaining hardware efficiency. Our contributions can be summarized as follows:


\begin{itemize}
    \item For the multiplication-free tiny neural network, we propose hybrid computing augmentation that leverages multiplicative operators to enhance the target multiplication-free network. While maintaining the same model structure, yields a more expressive and ultra-efficient network.

    \item A new weight sharing strategy is introduced for hybrid computing augmentation, which solves the weight discrepancy in heterogeneous (e.g., Gaussian vs. Laplacian) weight sharing during the augmentation process.

    \item Based on the idea of augmentation, a two-stage architecture search approach is adopted. Initially extract an augmented large network from the search space, followed by searching for deployable tiny networks within this augmented network.

\end{itemize}

\begin{figure*}[t]
    \centering
    \includegraphics[width=\linewidth]{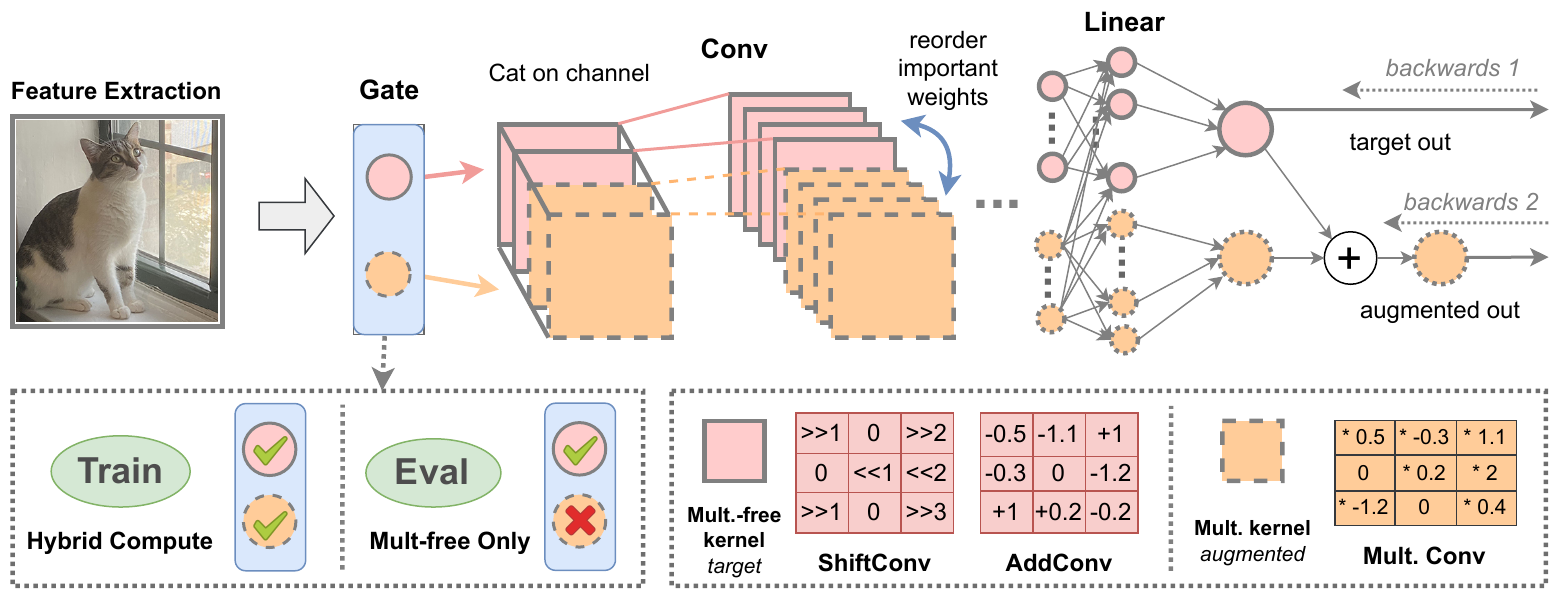}
    \vspace{-1.8em}
    \caption{ShiftAddAug augments weak operators with stronger ones. In this framework, pink indicates the multiplication-free kernels designated for the target model (either Shift or Add), while orange represents the multiplicative kernels of the augmented portion. During training, a wide weight is maintained, with the initial $n$ channels processed in a multiplication-free manner and the subsequent channels utilizing multiplicative operations. The weights of both are updated but only the multiplication-free part is exported for deployment. Therefore, the important weights in the wide weight will be reordered into the multiplication-free part. Obtained tiny models have higher accuracy (up to \hr{4.95\%}) than their directly trained counterparts.}
    \label{fig:overview}
    \vspace{-0.5em}
\end{figure*}

%% file: sec/2_rela.tex
\section{Related Works}


\indent 
\indent \textbf{Multiplication-Free NNs.} 
To mitigate the high energy and time costs associated with multiplication, a key strategy involves employing hardware-friendly operators instead of multiplications. 
ShiftNet\cite{wu2017shiftNet,chen2019shiftNet} proposes a zero-parameter, zero-flop convolution. DeepShift\cite{elhoushi2021deepshift} retains the calculation method of original convolution, but replaces the multiplication with bit-shift and bit-reversal. BNNs\cite{courbariaux2016BNN,lin2016BNN2,rastegari2016xnornet} binarize the weight or activation to build DNNs consisting of sign changes. AdderNet \cite{chen2021addernet,Song2021AdderSR} chooses to replace multiplicative convolution with less expensive addition, and design an efficient hardware implementation\cite{wang2021addernet_HWdesign}. ShiftAddNet\cite{you2020shiftaddnet} combines bit-shift and add, getting up to 196$\times$ energy savings on hardware 
as shown in Tab. \ref{tab:energy_saving}. 
ShiftAddVit\cite{you2023shiftaddvit} puts this idea into the vision transformer and performs hybrid computing through mixture of experts.

\vspace{-0.5em}
\input{tables/unit_energy}
\vspace{-1em}

\textbf{Network Augmentation.} 
Research on tiny neural networks are developing rapidly. Networks and optimization techniques designed for MCU have already appeared at present\cite{lin2020mcunet,lin2021mcunetV2}. Once-for-all\cite{Cai2020Once-for-All:} proposes the Progressive Shrinking and finds that the accuracy of the obtained model is better than the counterpart trained directly. Inspired by this result, NetAug\cite{NetAug} raises a point that tiny neural networks need more capacity rather than regularization in training. Therefore, they chose a scheme that is the opposite of regularization methods like Dropout\cite{Srivastava2014Dropout}: expand the model width and let the large model lead the small model to achieve better accuracy. 


\textbf{Neural Architecture Search.} 
NAS remarkably succeeded in automating the creation of efficient NN architectures\cite{liu2018darts, Liu2019Auto-DeepLab}, enhancing accuracy while incorporating hardware considerations like latency\cite{Tan2019MnasNet,Wu2019FBNet} and memory\cite{lin2020mcunet} usage into the design process.
NAS also extends its utility to exploring faster operator implementations\cite{chen2018tvm} and integrating network structures for optimization\cite{Lin2021NAAS,Shi2022NASA}, bringing designs closer to hardware requirements. ShiftAddNAS\cite{you2022shiftaddnas} pioneered a search space that includes both multiplicative and multiplication-free operators.

%% file: tables/unit_energy.tex

\begin{table}[h]
\caption{Energy cost (pJ) under 45nm CMOS.}
\vspace{-2em}
\vskip 0.15in
\begin{center}
\begin{small}
\begin{sc}
\begin{tabular}{lcccc}
\toprule
\textbf{OPs} & \textbf{FP32} & \textbf{FP16} & \textbf{INT32} & \textbf{INT8}\\
\midrule
        \textbf{Mult.} & 3.7 & 0.9  & 3.1  & 0.2   \\
        \textbf{Add} & 1.1  & 0.4 & 0.1 & 0.03   \\
        \textbf{Shift} & - & - & 0.13 & 0.024  \\
\bottomrule
\end{tabular}
\end{sc}
\end{small}
\end{center}
\vskip -0.1in
\label{tab:energy_saving}%
\end{table}

%% file: sec/3_method.tex
\section{ShiftAddAug}

\subsection{Preliminaries}

    \indent 
    \indent \textbf{Shift.}
    The calculation for the shift operator parallels that of standard linear or convolution operators using weight $W$, except that $W$ is rounded to the nearest power of 2. Bit-shift and bit-reversal techniques are employed to achieve calculation results equivalent to those obtained through traditional methods as Equ. \ref{equ:deepshift}. Inputs are quantized before computation and dequantized upon obtaining the output.

    \vspace{-1.0em}
    \begin{equation} \label{equ:deepshift}
    \begin{matrix}
            \left\{\begin{matrix}  
             S = \texttt{sign}( W) \\
             P = \texttt{round}(\log_2(\left |  W \right | ))
            \end{matrix}\right.
            \\
            \left\{\begin{matrix} 
             Y =  X {\tilde{ W_q}}^T =  X {( S \cdot  2^{ P} )}^T, \quad  train. \\
             Y = \sum_{i,j} \sum_{k}  \pm ( X_{i,k} <<  P_{k,j}),  \quad eval.
            \end{matrix}\right.
    \end{matrix}
    \end{equation}
    \vspace{-1.0em}
    
    \textbf{Add.}
    Add operator replaces multiplication with subtractions and $\ell_1$ distance since subtractions can be easily reduced to additions by using complement code.

    \vspace{-1.5em}
    \begin{equation} \label{equ:addConv}
         Y_{m,n,t}=-\sum_{i=0}^{d} \sum_{j=0}^{d} \sum_{k=0}^{c_{in}} \left |  X_{m+i,n+j,k}- F_{i,j,k,t} \right | .
    \end{equation}
    \vspace{-1.2em}
    
    \textbf{NetAug.} 
    Network Augmentation encourages the target tiny multiplicative NNs to work as a sub-model of a large model expanded in width. The target tiny NN and the augmented large model are jointly trained. The training loss and parameter updates are as follows:

    \vspace{-0.8em}
    \begin{equation} \label{equ:netaug}
    \begin{matrix}
        \mathcal L_{aug} = \mathcal L(W_t)+ \alpha \mathcal L(W_a), W_t \in W_a
        \\
        \\
        W_t^{n+1}=W_t^{n}-\eta(\frac{\partial \mathcal L(W_t^n)}{\partial W_t^n} + \alpha\frac{\partial \mathcal L(W_a^n)}{\partial W_t^n} ). 
    \end{matrix}
    \end{equation}
    \vspace{-0.8em}
    
    where $\mathcal L$ is the loss function, $W_t$ is the weight of the target tiny NN, $W_a$ is the weight of the augmented NN, and $W_t$ is a subset of $W_a$.

\subsection{Hybrid Computing Augment} \label{section:HCA}

    \indent 
    \indent ShiftAddAug goes a step further based on NetAug\cite{NetAug}, using strong operators to augment weak operators. 
    
    Taking an $n$-channel DepthWise Conv as an example, NetAug expands it by a factor of $\alpha$, resulting in a convolution weight for $\alpha n$ channels. During calculation, only the first $n$ channels are used for the target model, while the augmented model employs all $\alpha n$ channels. After trained as Equ. \ref{equ:netaug}, the important weights within $\alpha n$ channels are reordered into the first $n$, and only these $n$ channels are exported for deployment.
    
    As shown in Fig. \ref{fig:overview}, ShiftAddAug uses $[0, n)$ channels (target part) and $[n, \alpha n)$ channels (augmented part) for different calculation methods. The target part will use multiplication-free convolution ($\texttt{MFConv}$, ShiftConv\cite{elhoushi2021deepshift} or AddConv\cite{chen2021addernet} can be chosen) while multiplicative convolution ($\texttt{MConv}$, i.e. original Conv) is used as the augmented part. 


    Because the channels of Conv are widened, the \textit{input} of each convolution is also widened and can be conceptually split into the target part $X_t$ and the augmented part $X_a$, so does the \textit{output} $Y_t,Y_a$. In ShiftAddAug, $X_t$ and $Y_t$ mainly carry information of $\texttt{MFConv}$, while $X_A$ and $Y_A$ are obtained by original Conv.  

    Here three types of operators commonly used to build tiny NNs are discussed: Convolution (Conv), DepthWise Convolution (DWConv), and Fully Connected (FC) layer. The hybrid computing augmentation for DWConv is the most intuitive: split the input into $X_t$ and $X_a$, then use $\texttt{MFConv}$ and $\texttt{MConv}$ to calculate respectively and connect the obtained $Y_t$ and $Y_a$ in the channel dimension. For Conv, We use all input $X$ to get $Y_a$ through $\texttt{MConv}$. But to get $Y_t$, we still need to split the input and calculate it separately, and finally add the results. Since the FC layer is only used as the classification head, its output does not require augmentation. We divide the input and use $Linear$ and $ShiftLinear$ to calculate respectively, and add the results. If bias is used, it will be preferentially bounded to multiplication-free operators.

    \vspace{-1.0em}
    \begin{equation} \label{equ:shiftaddaug}
        \begin{matrix}

        DWConv: \left\{\begin{matrix}
         Y_t = \texttt{MFConv}(X_t)\\
         Y_a = \texttt{MConv}(X_a)\\
         Y = \texttt{cat}(Y_t, Y_a)
        \end{matrix}\right.
        \\
        \\
        Conv: \left\{\begin{matrix}
         Y_t = \texttt{MFConv}(X_t) + \texttt{MConv}(X_a)\\
         Y_a = \texttt{MConv}(X)\\
         Y = \texttt{cat}(Y_t, Y_a)
        \end{matrix}\right. 
          \\
          \\
        FC: \left\{\begin{matrix}
         Y_t = \texttt{ShiftLinear}(X_t)\\
         Y_a = \texttt{Linear}(X_a)\\
         Y = Y_t + Y_a 
        \end{matrix}\right.

        \end{matrix}
    \end{equation}
    \vspace{-0.8em}

\subsection{Heterogeneous Weight Sharing} \label{section:HWS}

    \vspace{-0.5em}
    \begin{figure}[h]
        \centering
        \includegraphics[width=\linewidth]{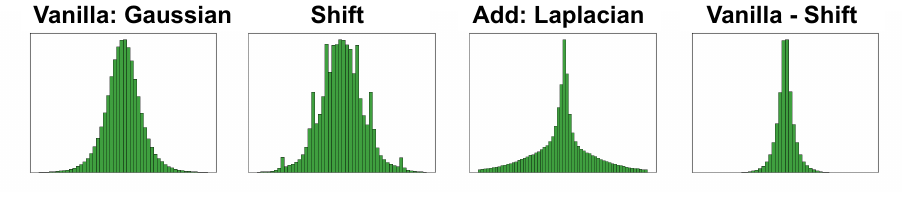}
        \vspace{-2.5em}
        \caption{Weight distribution of different convolution operators in MobileNetV2-w0.35. Inconsistent weight distribution leads to discrepancy, making weight sharing difficult.}
        \label{fig:weightDist}
        \vspace{-0.5em}
    \end{figure}
    
    \indent
    \indent\textbf{Dilemma.} 
    The important weights will be reordered to the target part at each end of the training epoch ($\ell_1$ norm for importance). This is a process of weight sharing and is the key to effective augmentation.

    However, the weight distribution of the multiplication-free operator is inconsistent with the original Conv. It causes the weight discrepancy, i.e. \textit{good weights in original Conv may not be good in MFConv}. As shown in Fig. \ref{fig:weightDist}, the weight in the original Conv conforms to Gaussian distribution, while ShiftConv has spikes at some special values. The weight in AddConv conforms to the Laplace distribution. The weight in ShiftConv is the one of the original Conv plus a Laplace distribution with a small variance. 
    
    ShiftAddNas\cite{you2022shiftaddnas} adds a penalty term to the loss function, guides the weight to conform to the same distribution. It affects the network to achieve its maximum performance. The Transformation Kernel they proposed also doesn't work on our approach since the loss diverges as Tab. \ref{tab:convergence}. We argue that their approach makes training unstable. This dilemma motivated us to propose our heterogeneous weight sharing strategy.

    \begin{figure}[t]
        \centering
        \includegraphics[width=\linewidth]{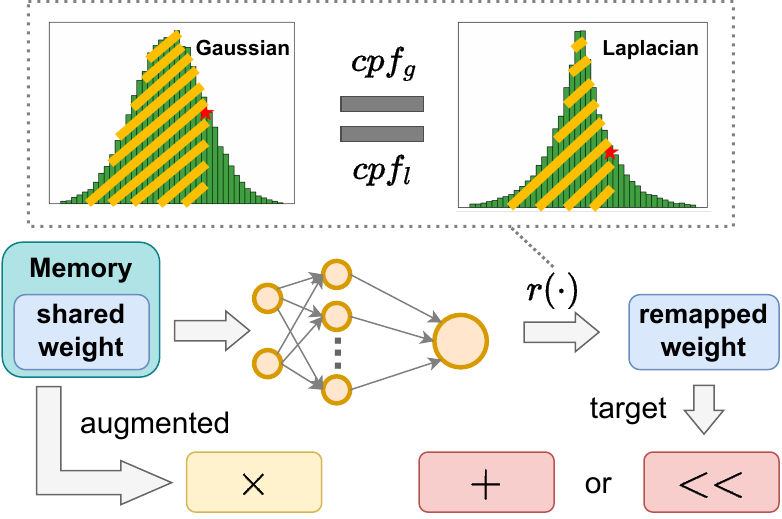}
        \vspace{-2.0em}
        \caption{Weight remapping strategy creates mappings between different weight distributions, making weight sharing workable. $cpf_g$ for cumulative probability function (CPF) for Gaussian distribution and $cpf_l$ CPF for Laplace distribution}
        \label{fig:weightRemap}
        \vspace{-0.5em}
    \end{figure}

    \textbf{Solution: heterogeneous weight sharing.}
    To solve the dilemma above, we propose a new heterogeneous weight sharing strategy for the shift and add operators. This method is based on original Conv and remap parameters to weights of different distribution through a mapping function $\mathcal{R}(\cdot )$. In this way, all weights in memory will be shared under the Gaussian distribution, but will be remapped to an appropriate state for calculation.

    When mapping the Gaussian distribution to the Laplace distribution, we hope that the cumulative probability of the original value and mapping result is the same. Firstly, calculate the cumulative probability of the original weight in Gaussian. Then put the result in the percent point function of Laplacian. The workflow is shown in Fig. \ref{fig:weightRemap}. The mean and standard deviation of the Gaussian can be calculated through the weights, but for the Laplace, these two values need to be determined through prior knowledge.

    \vspace{-1.0em}
    {\setlength\belowdisplayskip{5pt}
    \begin{equation}
        \label{equ:weightRemap}
        \begin{aligned}
        W_l &= \mathcal{R}(W_g) = r(\texttt{FC}(W_g))  \\
        r(\cdot ) &= \texttt{ppf}_l(\texttt{cpf}_g(\cdot))    \\ 
        \texttt{cpf}_g(x) &= \frac{1}{\sigma \sqrt{2\pi}} \int_{-\infty }^{x} e^{(-\frac{(x-u)^2}{2\sigma^2} )} \mathrm{d}x\\
        \texttt{ppf}_l(x) &= u - b *\texttt{sign}(x-\frac{1}{2})*\ln (1-2\left | x-\frac{1}{2} \right | )
        \end{aligned}
    \end{equation}}

    Where $W_g$ is the weight in original Conv that conforms to the Gaussian distribution, and $W_l$ is the weight obtained by mapping that conforms to the Laplace distribution. $\texttt{FC}$ is a fully connected layer, which is previously trained and frozen in augmented training. We need this because the weights don't fit the distribution perfectly. $\texttt{cpf}_g(\cdot)$ is the cumulative probability function of Gaussian, $\texttt{ppf}_l(\cdot)$ is the percentage point function of Laplace.

\subsection{Neural Architecture Search}

    \begin{figure*}[t]
        \centering
        \includegraphics[width=1.0\linewidth]{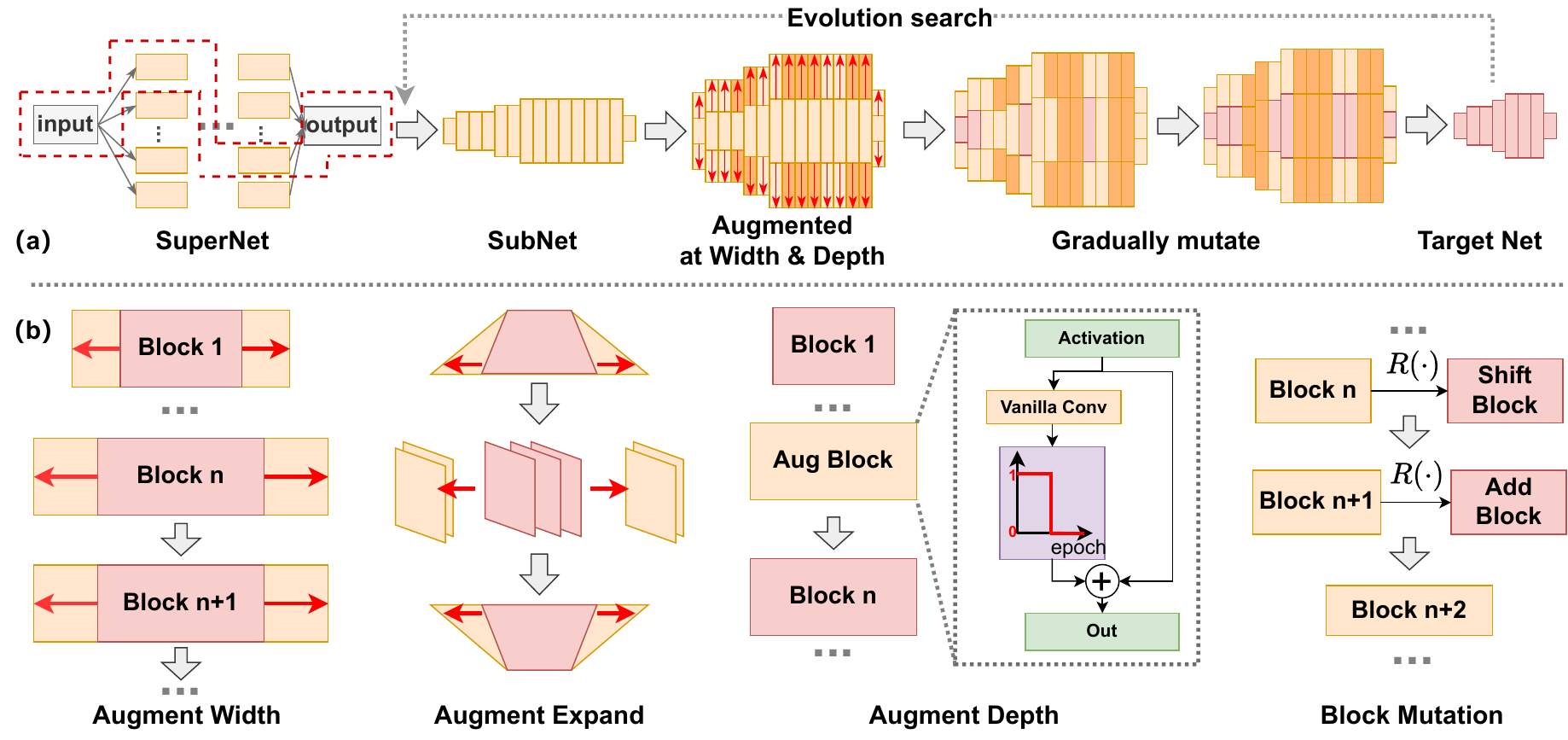}
        \vspace{-2.0em}
        \caption{Light orange block for $\texttt{MConv}$; pink block for $\texttt{MFConv}$; deep orange block for Depth Augmentation.  \textbf{(a).}Search process: two-stage search. Find a SubNet as depth augmented one and then further cut out a tiny TargetNet on it for development. Start with multiplication and gradually convert the TargetNet to multiplication-free during training. \textbf{(b).}Methods to augment. \textit{Augment Width:} use $\texttt{MConv}$ to widen the $\texttt{MFConv}$ channel; \textit{Augment Expand:} increase expand ratio of depthwise separable convolution; \textit{Augment Depth:} Augmented blocks for depth. Only participate in training and will not be exported; \textit{Block Mutation:} start with $\texttt{MConv}$, mutate the block into $\texttt{MFConv}$ during training.}
        \label{fig:NAS}
        \vspace{-1.0em}
    \end{figure*}

    \indent 
    \indent To obtain SOTA multiplication-free model at the tiny model size, a two-stage NAS is proposed.

    Based on the idea of augmentation, ShiftAddAug starts from a multiplicative SuperNet and cuts a deep SubNet from it as depth-augmented NN. Then select some layers on the SubNet to form the tiny TargetNet for final use. The TargetNet should meet the pre-set hardware limitation. This setup allows the TargetNet to be a part of the SubNet, facilitating joint training with weight sharing as Equ. \ref{equ:netaug}. Layers within the SubNet that are not selected for the TargetNet serve as a form of depth augmentation. Moreover, the layers used for deep augmentation are initially selected but gradually phased out from the target network in training progresses.

    A new block mutation training is also proposed, which tends to gradually transform multiplication operators into multiplication-free states during training to make the training process more stable. The training process starts with all multiplications, and the layers of the target network become multiplication-free one by one from shallow to deep. At the end of training, a completely multiplication-free TargetNet can be obtained.

    While ShiftAddNas\cite{you2022shiftaddnas} directly uses hybrid computing to train their SuperNet and directly cut SubNets that meet the hardware requirements, we start from the multiplicative SuperNet and split the search process into two steps. The middle step is used for augmented training, which is the unique part of ShiftAddAug.

    \input{tables/augmented_searchSpace}

    Combining the Width Augmentation and Expand Augmentation we used in section \ref{section:HCA}, we construct our search space for the augmentation part according to Tab. \ref{tab:augmented_searchSpace}.
    
    We follow tinyNAS\cite{lin2020mcunet} to build SuperNet and cut SubNet. Then use evolutionary search to search for subsequent steps.

%% file: tables/augmented_searchSpace.tex

\vspace{-0.2em}
\begin{table}[t]
\caption{Search space for Augmentation.}
\vspace{-2em}
\label{sample-table}
\vskip 0.15in
\begin{center}
\begin{small}
\begin{sc}
\begin{tabular}{cccc}
    \toprule
     \textbf{Item}  & \textbf{Optional} \\
    \midrule
     SubNet as aug. model &  SuperNet \\
     TargerNet & SubNet \\
     Block types & [Mult. , Shift, Add] \\
     width aug. multiples  & [2.2, 2.4, 2.8, 3.2] \\
     expand aug. multiples  & [2.2, 2.4, 2.8, 3.2] \\
     Time node: mutation start  & [5\%, 10\%, 15\%, 20\%] \\
     Time node: mutation stop  & [50\%, 60\%, 70\%] \\
    \bottomrule
\end{tabular}
\end{sc}
\end{small}
\end{center}
\vskip -0.1in
\label{tab:augmented_searchSpace}%
\end{table}

%% file: sec/4_exp.tex
\section{Experiments}

\subsection{Setup}

    \indent 
    \indent \textbf{Datasets.}
    We conduct experiments on several image classification datasets, including CIFAR10\cite{Krizhevsky2009CIFAR}, CIFAR100\cite{Krizhevsky2009CIFAR}, ImageNet-1K\cite{Deng2009ImageNet}, Food101\cite{Bossard2014Food101}, Flowers102 \cite{Nilsback2008Flower102}, Cars\cite{Krause2013Cars}, Pets\cite{Parkhi2012Pets}. We also evaluated our method on VOC\cite{Mark2010VOC} and OpenEDS\cite{palmero2020openeds2020} for segmentation task.
    
    \textbf{Training Details.}
    The NNs are trained with batch size 128 using 2 GPUs. The SGD optimizer is used with Nesterov momentum 0.9 and weight decay 4e-5. By default, the Mult. and Shift models are trained for 250 epochs and Add models are trained for 300 epochs. The initial learning rate is 0.05 and gradually decreases to 0 following the cosine schedule. Label smoothing is used with a factor of 0.1. For ShiftConv and ShiftLinear, the weight is quantized to 5bit Power-of-2 and the activation is quantized to 16bit. AddConv is calculated under 32bit.
    
    \textbf{Hardware Performance.}
    Since many works have verified the efficiency of shift and add on proprietary hardware\cite{you2020shiftaddnet,you2022shiftaddnas,wang2021addernet_HWdesign,you2023shiftaddvit}, we follow their evaluation metrics. Hardware energy and latency are measured based on a simulator of Eyeriss-like hardware accelerator\cite{Chen2017Eyeriss1,Zhao2020Eyeriss2}, which calculates not only computational but also data movement energy.

\input{tables/shiftaddaugVSbaseline}
    \input{tables/moreClassifyDatasets}

\subsection{ShiftAddAug vs. Baseline}

    \indent 
    \indent We validate our method on MobileNetV2\cite{sandler2019mobilenetv2}, MobileNetV3\cite{howard2019mobilenetv3}, MCUNet\cite{lin2020mcunet}, ProxylessNAS\cite{cai2019proxylessnas}, MobileNetV2-Tiny\cite{lin2020mcunet}. ShiftAddAug provides consistent accuracy improvements (average \hrup{2.82\%}) for ShiftConv augmentation over the multiplicative baselines. For AddConv augmentation, it improves the accuracy compared with direct training (average \hrup{1.59\%}).  The resulting model will be faster (3.0$\times$ for Shift) and more energy-efficient (\hrdown{68.58\%} for Shift and \hrdown{52.02\%} for Add) due to the use of hardware-friendly operators. As shown in Tab. \ref{tab:CompareNetAug}, these multiplication-free operators usually hurt the performance of the network. Changing all operators to Shift will cause \hrdown{0.82\%} accuracy drop on average compared to the multiplication baseline. But after using our method, the accuracy increased by \hrup{3.63\%} on average under the same energy cost. \footnote{Loss becomes NaN when we use AddConv on MobileNetV3, both direct training and augmented training.}
    
    In addition, our method achieves higher results than multiplicative NetAug on some models (MobileNetV3:\hrup{1.17\%}, MCUNet:\hrup{1.44\%}, ProxylessNAS:\hrup{1.54\%}). This means that our method enables the multiplication-free operator to be stronger than those of the original operator. 
    
    To verify the generality of our method, we also conduct experiments on more datasets. As shown in Tab. \ref{tab:MoreDatasets}, our method can achieve \hrup{0.89\%} to \hrup{4.28\%} accuracy improvements on different datasets. Hybrid computing augmentation works better on smaller models and datasets with less classification. On Flower102, MobileNetV2-w0.35 has \hrup{3.83\%} accuracy improvements with our method, while MCUNet has only \hrup{1.47\%}. This shows that a smaller model capacity can achieve a better effect on this dataset. The larger the model, the smaller the gain brought by augmentation. The same phenomenon also occurs in CIFAR10. For bigger datasets such as ImageNet, even if it is augmented, the capacity of the model is still not enough. It achieves \hrup{1.94\%} for MobileNetV2-w0.35 and \hrup{0.89\%} for MCUNet on ImageNet. But this goes beyond the common work scenarios of IOT devices.

\subsection{ShiftAddAug vs. SOTA Mult.-Free Models}

    \begin{figure}[h]
        \centering
        \includegraphics[width=0.75\linewidth]{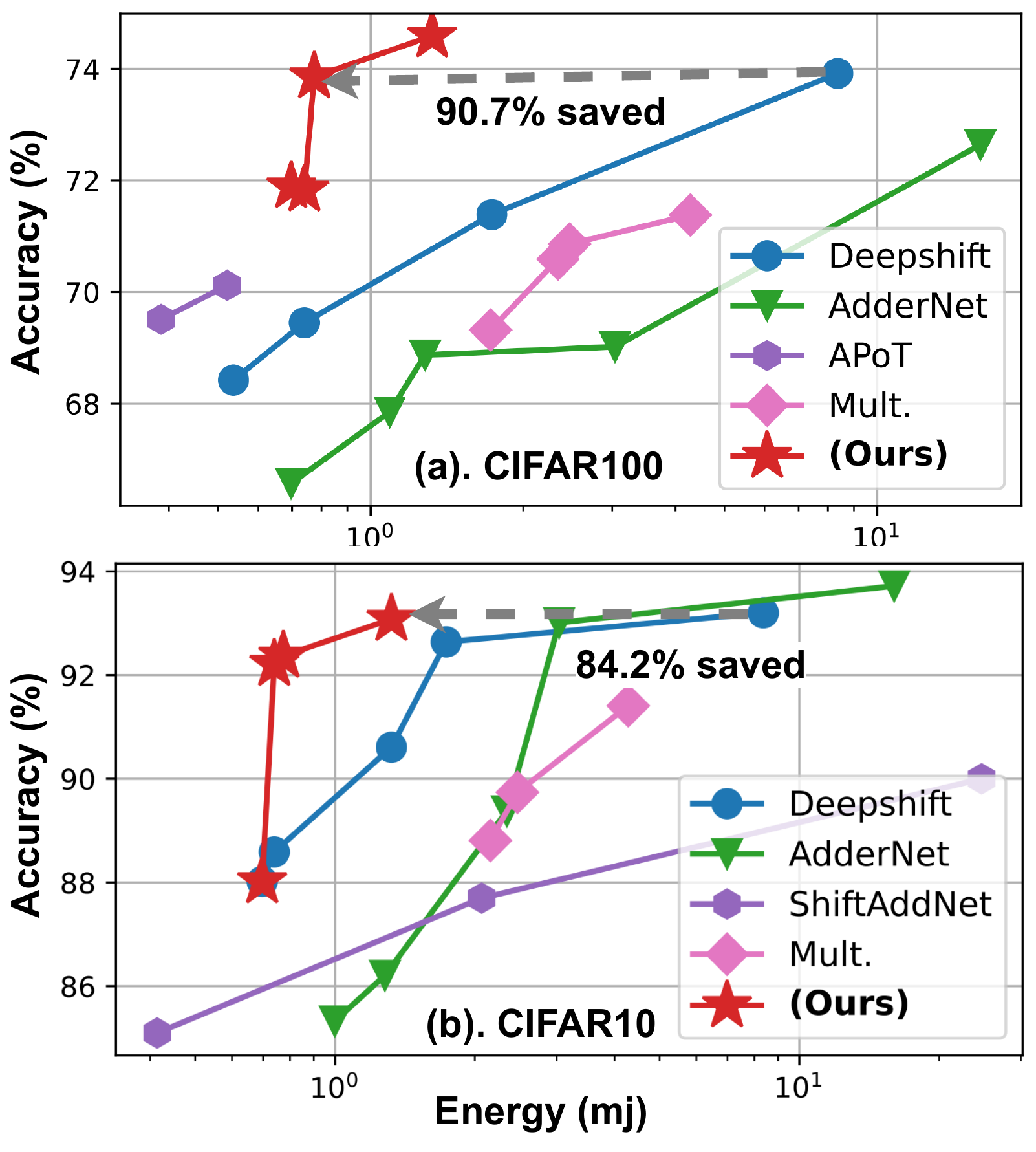}
        \vspace{-0.5em}
        \caption{Accuracy and energy cost of ShiftAddAug over SOTA manually designed multiplication-free model and tiny multiplicative models. Tested on CIFAR-100/10.}
        \label{fig:previsou}
        \vspace{-1.5em}
    \end{figure}
    
    \indent 
    \indent We compare ShiftAddAug over SOTA multiplication-free models, which are designed manually for tiny computing devices, on CIFAR-10/100 to evaluate its effectiveness. As shown in Fig. \ref{fig:previsou}, the model structures we use are smaller and have better energy performance. With ShiftAddAug, the accuracy still exceeds existing work. For DeepShift and AdderNet, our method boosts \hrup{0.67\%} and \hrup{1.95\%} accuracy on CIFAR100 with \hrdown{84.17\%} and \hrdown{91.7\%} energy saving. Compared with the SOTA shift quantization method APoT\cite{Li2020APOT}, we achieve an improved accuracy of \hrup{3.8\%}. With the same accuracy on CIFAR10, our model saves \hrdown{84.2\%} of the energy compared with Deepshift, and \hrdown{56.45\%} of the energy compared with AdderNet.

\subsection{ShiftAddAug on Segmentation Task}
    \indent 
    \indent To further verify the effectiveness of our method, we conduct experiments on the semantic segmentation task. We choose VOC\cite{Mark2010VOC} for general performance and OpenEDS\cite{palmero2020openeds2020} for specific tasks on MCU level devices. As shown in Tab. \ref{tab:segmentation},  ShiftAddAug has greater improvement on OpenEDS, which illustrates the application potential of our method. 

\begin{figure}[h]
        \centering
        \includegraphics[width=1.0\linewidth]{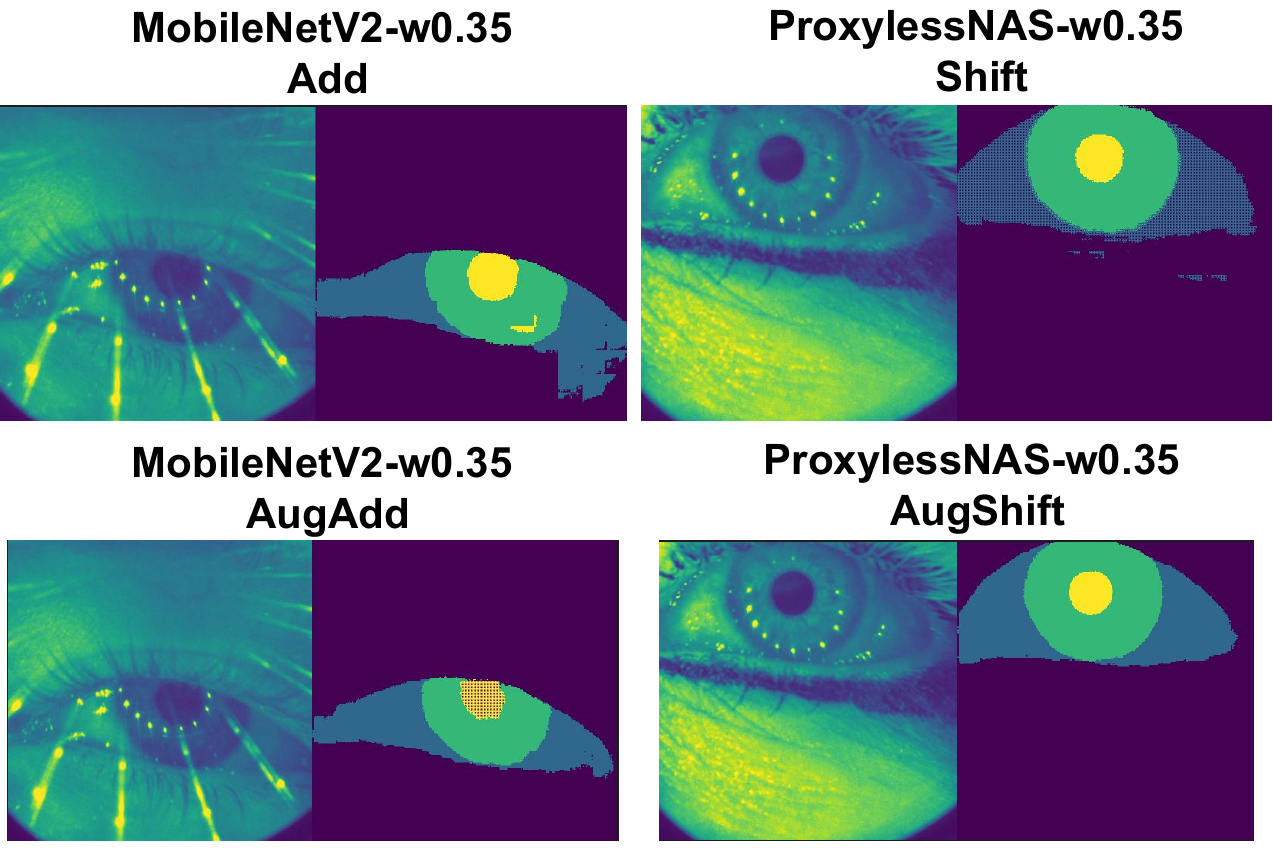}
        \vspace{-1.0em}
        \caption{ShiftAddAug on OpenEDS.}
        \label{fig:iris}
        \vspace{-1.0em}
    \end{figure}

    From the results shown in Fig.\ref{fig:iris}, we can see that the model trained with augmentation will have fewer abnormal segmentation areas.
    \input{tables/segmentation}

\subsection{ShiftAddAug on Neural Architecture Search}

    \input{tables/nas}

    \indent 
    \indent Based on hybrid computing augmentation, we introduce neural architecture search into our method to get stronger tiny neural networks. We conduct our experiments on Cifar-10/100 and compare them with the results of ShiftAddNAS\cite{you2022shiftaddnas} which is better than multiplication-based FBNet\cite{Wu2019FBNet}. As shown in Tab. \ref{tab:nas}, the multiplication-free model we obtained achieved higher accuracy (\hrup{3.61\%}) than ShiftAddNas. For hybrid-computed models, we have to use a smaller input resolution (96 instead of 160) and larger models. While the input resolution of ShiftAddNas is 32, this would give us 9$\times$ the number of calculations if we have the same model architecture. Even so, we can still save 37.1\% of calculations on average with similar accuracy.

\subsection{Ablation Study}
    \begin{figure*}[t]
        \centering
        \includegraphics[width=0.75\linewidth]{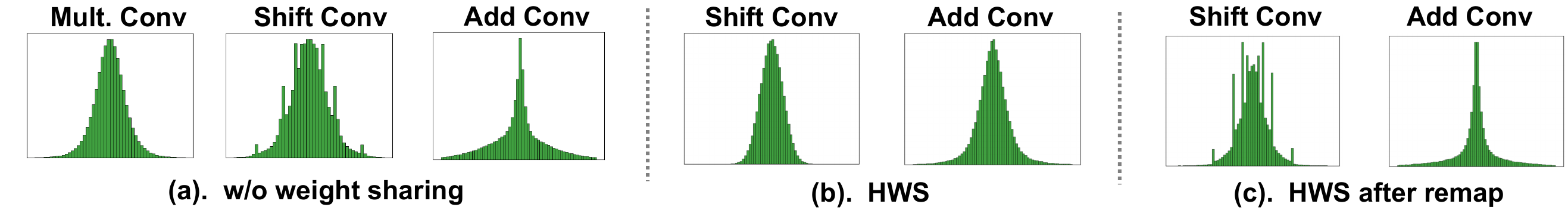}
        \vspace{-1.0em}
        \caption{The weight distribution of original Conv / ShiftConv / AddConv layers}
        \label{fig:ablation_HWS}
        \vspace{-0.5em}
    \end{figure*}
    
    \indent 
    \indent \textbf{Hybrid Computing Augment.} 
    In order to prove that hybrid computing works better, we add an experiment using only multiplication-free operators for augmentation. We exert experiments based on NetAug, and replace all the original operators with Shift operators. The difference from our method is that the Shift operator is also used in the augmentation part, while our method uses multiplication in it. As shown in Tab. \ref{tab:ablation_shiftaug}, it yields an average accuracy improvement of \hrup{1.02\%}. 
    
    Then without using the heterogeneous weight sharing (HWS) method, just augmenting tiny NNs with the multiplicative operator will cause \hrdown{0.9\%} accuracy drop on average due to the weight tearing problem. However, the situation changed after we applied for HWS. Compared with using the shift operator for augmentation, the accuracy increased by \hrup{2.2\%}.
    \input{tables/ablation_shitaddaug}

    \textbf{Heterogeneous Weight Sharing.} 
    Since the help of HWS on training results has been discussed above, here we visualize the weight distributions of Conv layers in tiny NNs under three scenarios, (a) w/o weight sharing; (b) heterogeneous weight sharing; (c) weight after remapped, as shown in Fig. \ref{fig:ablation_HWS}. We consistently observe that the three operators exhibit different weight distributions without weight sharing. With our HWS, the saved weights are based on the original Conv and conform to Gaussian distribution. After remapping, the weights can show different distribution states in Shift/Add Conv calculations.     

    Different methods for solving weight tearing are also compared. As shown in Tab. \ref{tab:convergence}, both the ShiftAddNas method and direct linear remapping will cause loss to diverge. The additional penalty has a larger impact on tiny NNs, causing accuracy to decrease.
    
    Additionally, Tab. \ref{tab:rebuttal_weightsharing} shows why weight sharing is important, which also explains why we have to face the problem of heterogeneous weights. Tab. \ref{tab:rebuttal_woAug} tells that HWS is not a parameterized trick directly improving accuracy, but just a compensation in augmentation. 
    
    \input{tables/convergence}
    \input{tables/rebuttal_weightshare}
    \input{tables/rebuttal_woAug}

    \textbf{Neural Architecture Search.}
    Our neural architecture search approach is dependent on our proposed hybrid computing augmentation. And it can help the multiplication-free operator to be as strong as the original operator. Block augmentation and block mutation help us further improve the performance of multiplication-free tiny NNs. As shown in Tab. \ref{tab:ablation_nas}, under similar energy consumption and latency, block augmentation improves accuracy by \hrup{0.5\%}, and block mutation improves by \hrup{0.79\%}. Combining all methods, the accuracy of the target model is increased by \hrup{1.22\%}.
    
    \input{tables/ablation_nas}

    \input{tables/moreClassifyDatasets_moreResults}

    \subsection{Limitation}

    \indent 
    \indent Since we use additional multiplication structures for augmentation, training consumes more resources. This is consistent with NetAug. 
    
    In terms of memory usage, which depends on the size of the augmented NN, ShiftAddAug usually doubles or triples it compared with direct training. This makes ShiftAddAug not suitable for on-device training.

%% file: tables/shiftaddaugVSbaseline.tex
\begin{table*}[htbp]
  \setlength{\tabcolsep}{4pt}
  \centering
  \caption{ShiftAddAug vs. Multiplicative and directly trained Multiplication-free Baseline in terms of accuracy and efficiency on CIFAR100 classification tasks. The results obtained by our method are bolded. The accuracy on the left is directly trained and the one on the right is obtained with augmentation. \textit{Base} for Mult.; \textit{Shift} for ShiftConv in DeepShift and \textit{Add} for AddConv in AdderNet.}
  \vspace{-1em}
  \renewcommand{\arraystretch}{1.15} 
  \resizebox{\linewidth}{!}{
    \begin{tabular}{c|c|cccc|ccc}
    \hline
    \hline
    \textbf{Model} & \textbf{Method} & \textbf{Params (M)} & \textbf{Mult (M)} & \textbf{Shift (M)} & \textbf{Add (M)} & \textbf{Accuracy(\%)} & \textbf{Energy (mj)} & \textbf{Latency (ms)}  \\
    \hline
    \hline
    \multirow{2}[0]{*}{MobileNetV2}
        & Base / NetAug              & 0.52 &29.72 & 0 & 29.72 & 70.59 / 71.98 & 2.345 & 0.73 \\
    \multirow{2}[1]{*}{w0.35}
        & Shift / \textbf{AugShift}  & 0.52 &0 & 29.72 & 29.72 & 69.25 / \textbf{71.83} (\up{2.58}) & 0.74 & 0.246 \\
        & Add / \textbf{AugAdd}      & 0.52 & 4.52 & 0 & 56.88 & 67.85 / \textbf{69.38} (\up{1.5}) & 1.091 & 0.753 \\
        
    \hline
    
    \multirow{2}[0]{*}{MobileNetV3}
        & Base / NetAug              & 0.96 &18.35 & 0 & 18.35 & 69.32 / 72.2 & 1.726 & 0.485  \\
    \multirow{2}[1]{*}{w0.35}
        & Shift / \textbf{AugShift}  & 0.96 &0 & 18.35 & 18.35 & 68.42 / \textbf{73.37} (\up{4.95}) & 0.536 & 0.16 \\
        & Add / \textbf{AugAdd}      & 0.96 & 3.5 & 0 & 34.34 & - / - & 0.699 & 0.512 \\
        
    \hline
        \multirow{3}[0]{*}{MCUNet - in3}
        & Base / NetAug              & 0.59 &65.72 & 0 & 65.72 & 71.38 / 73.15 & 4.28 & 1.682 \\
        & Shift / \textbf{AugShift}  & 0.59 &0 & 65.72 & 65.72 & 70.87 / \textbf{74.59} (\up{3.72}) & 1.323 & 0.545 \\
        & Add / \textbf{AugAdd}      & 0.59 & 20.91 & 0 & 113.09 & 70.25 / \textbf{72.72} (\up{2.47}) & 2.345 & 1.72 \\
        
    \hline

    \multirow{2}[0]{*}{ProxylessNAS}
        & Base / NetAug              & 0.63 &34.56 & 0 & 34.56 & 70.86 / 72.32 & 2.471 & 0.883 \\
    \multirow{2}[1]{*}{w0.35}
        & Shift / \textbf{AugShift}  & 0.63 &0 & 34.56 & 34.56 & 70.54 / \textbf{73.86} (\up{3.32}) & 0.774 & 0.294 \\
        & Add / \textbf{AugAdd}      & 0.63 & 8.81 & 0 & 61.97 & 68.87 / \textbf{70.18} (\up{1.31}) & 1.281 & 0.881 \\
        
    \hline
    
    \multirow{2}[0]{*}{MobileNetV2}
        & Base / NetAug              & 0.35 &27.31 & 0 & 27.31 & 69.3 / 71.62 & 2.161 & 0.67  \\
    \multirow{2}[1]{*}{-Tiny}
        & Shift / \textbf{AugShift}  & 0.35 &0 & 27.31 & 27.31 & 68.29 / \textbf{71.89} (\up{3.6}) & 0.697 & 0.228 \\
        & Add / \textbf{AugAdd}      & 0.35 & 4.43 & 0 & 52.09 & 66.57 / \textbf{67.65} (\up{1.08}) & 0.999 & 0.693 \\
    
    \hline
    \hline
    \end{tabular}%
    }
  \label{tab:CompareNetAug}%
\end{table*}%

%% file: tables/moreClassifyDatasets.tex
\begin{table*}[t]
  \setlength{\tabcolsep}{4pt}
  \centering
  \caption{Accuracy of MobileNetV2 (w0.35) and MCUNet-in3 on more datasets. ShiftAddAug can improve performance without any inference overhead on fine-grained classification tasks. Please refer to Tab.\ref{tab:MoreDatasets_moreResults} for results on more models}
  \vspace{-1em}
  \renewcommand{\arraystretch}{1.2} 
  \resizebox{0.8\linewidth}{!}{
    \begin{tabular}{c|c|c|c|c|c|c|c}
    \hline
    \hline
    \multirow{2}[1]{*}{\textbf{Model}} & \multirow{2}[1]{*}{\textbf{Methods}} & \multirow{2}[1]{*}{\textbf{\space CIFAR10 \space }} & \multirow{2}[1]{*}{\textbf{ \space ImageNet \space }} & \multirow{2}[1]{*}{\textbf{ \space Food101 \space   }} & \multirow{2}[1]{*}{\textbf{\space Flower102 \space }} & \multirow{2}[1]{*}{\textbf{\quad Cars \quad}} & \multirow{2}[1]{*}{\textbf{ \quad Pets  \quad }} \\
      & & & & & & & \\
    \hline
    \hline 
    \multirow{2}[1]{*}{\textbf{MobileNetV2 - w0.35}}
        & Shift               & 88.59  & 51.92 & 72.99 & 92.25  & 72.83  & 75.4  \\
        & \textbf{AugShift}   & 92.51  & 53.86 & 74.67 & 96.08  & 74.47  & 79.59  \\
    \hline 
    \multirow{2}[2]{*}{\textbf{MCUNet - in3}}
        & Shift              & 90.61  & 56.45 & 78.46 & 95.59  & 80.51  & 79.67  \\
        & \textbf{AugShift}  & 93.08  & 57.34 & 79.96 & 97.06  & 83.29  & 83.95  \\
    \hline
    \hline
    \end{tabular}%
  }
  \label{tab:MoreDatasets}%
  \vspace{-0.5em}
\end{table*}%

%% file: tables/segmentation.tex
\begin{table}[t]
  \setlength{\tabcolsep}{4pt}
  \centering
  \caption{ShiftAddAug on semantic segmentation task (mIoU \%)}
  \vspace{-1em}
  \renewcommand{\arraystretch}{1.25} 
  \resizebox{\linewidth}{!}{
    \begin{tabular}{l|c|c|c}
    \hline
    \hline
    \multirow{2}[1]{*}{\textbf{Dataset}} & \multirow{2}[1]{*}{\textbf{Method}}
 & \textbf{MobileNetV2} & \textbf{ProxylessNAS} \\
    & & \textbf{w0.35} & \textbf{w0.35} \\
    \hline
    \hline

    \multirow{2}[1]{*}{VOC} & Base/NetAug & 66.23 / 69.72 & 66.63 / 68.09 \\
    &Shift/\textbf{AugShift} & 66.02 / \textbf{68.78} & 65.8 / \textbf{68.98} \\

    \hline
    
    \multirow{3}[1]{*}{OpenEDS} & Base/NetAug & 88.68 / 92.41 & 86.27 / 92.84 \\
    &Shift/\textbf{AugShift} & 88.01 / \textbf{94.52} & 86.01 / \textbf{95.12} \\
    &Add/\textbf{AugAdd} & 83.94 / \textbf{91.20} & 82.01 / \textbf{90.45} \\
        
    \hline
    \hline
    \end{tabular}%
    }
  \label{tab:segmentation}%
\end{table}%

%% file: tables/nas.tex
\begin{table*}[t]
  \setlength{\tabcolsep}{4pt}
  \centering
  \caption{ShiftAddAug vs. SOTA NAS method for hybrid operators in term of accuracy and efficiency on Cifar-10/100 classification tasks. ShiftAddAug can further improve the performance of the multiplication-less model.}
  \vspace{-1em}
  \renewcommand{\arraystretch}{1.15} 
  \resizebox{0.8\linewidth}{!}{
    \begin{tabular}{c|cc|ccc|cc}
    \hline
    \hline
    \multirow{2}[0]{*}{\textbf{Dataset}} & \multirow{2}[0]{*}{\textbf{Method}} & \multirow{2}[0]{*}{\textbf{Res.}} & \multicolumn{3}{c|}{\textbf{Calculation Amount}} & \multirow{2}[0]{*}{\textbf{Accuracy(\%)}} & \multirow{2}[0]{*}{\textbf{MACs Compare}} \\
    
     &  &  & \textbf{Mult (M)} &\textbf{Shift (M)} & \textbf{Add (M)} &  &  \\
    
    \hline
    \hline
    \multirow{3}[0]{*}{CIFAR10}
        & ACGhostNet-B\cite{Li2022ACGhost}              & 32 & 28 & 0 &61 & 95.2  & 1.62$\times$   \\
        & ShiftAddNas              & 32 &17 & 19 & 58 & 95.83 & 1.71$\times$  \\
        & \textbf{ShiftAddAug}             & 96  &12.3 & 14.5 & 28 & \textbf{95.92} & 1.0$\times$ \\
    \hline
    \multirow{4}[2]{*}{CIFAR100}
        & ShiftAddNas (Mult-free)  & 32    &3 & 35 & 48 & 71.0 & 1.27$\times$   \\  
        & ShiftAddNas      & 32           &22 & 21 & 62 & 78.6 & 1.44$\times$  \\
        & \textbf{ShiftAddAug (Mult-free)}  & 160 & 0.13 & 33.5 & 33.6 & \textbf{74.61} & 1.0$\times$ \\
        & \textbf{ShiftAddAug} & 96  &16.2 & 20.2 & 36.4 & \textbf{78.72} & 1.0$\times$  \\
    \hline
    \hline
    \end{tabular}
    }
  \label{tab:nas}
  \vspace{-0.5em}
\end{table*}

%% file: tables/ablation_shitaddaug.tex
    





\begin{table}[h]
  \setlength{\tabcolsep}{4pt}
  \centering
  \caption{The ablation study of hybrid computing augmentation and heterogeneous weight sharing in terms of accuracy on CIFAR100.}
  \vspace{-1em}
  \renewcommand{\arraystretch}{1.2} 
  \resizebox{\linewidth}{!}{
    \begin{tabular}{c|c|c|c}
    \hline
    \hline
    \multirow{2}[1]{*}{\textbf{Method}} & \multirow{1}[0]{*}{\textbf{MobileNetV2}} & \multirow{1}[0]{*}{\textbf{MCUNet}} & \multirow{1}[0]{*}{\textbf{ProxylessNAS}} \\
      & \multirow{1}[1]{*}{\textbf{w0.35}} & \multirow{1}[1]{*}{\textbf{in3}} & \multirow{1}[1]{*}{\textbf{w0.35}} \\
    \hline
    \hline 
    Mult. baseline & 70.59&	71.38&	70.86 \\
    
    To shift op & 69.25& 70.87&	70.54	\\

    Aug. with Shift & 70.12& 72.68&	70.91	 \\

    Aug. with Hybrid Computation & 69.41& 71.02&	70.6 \\

    \hline

    \textbf{Aug. with HWS} & 71.83&	74.59&	73.86\\
    \hline
    \hline
    \end{tabular}%
  }
  \label{tab:ablation_shiftaug}%
\end{table}%

%% file: tables/convergence.tex
\begin{table}[h]
  \setlength{\tabcolsep}{4pt}
  \caption{The ablation study of different method for HWS}
  \centering
  \vspace{-1em}
  \renewcommand{\arraystretch}{1.25} 
  \resizebox{\linewidth}{!}{
    \begin{tabular}{c|c|c|c}
    \hline
    \hline
    \textbf{Method} & \textbf{MobileNetV2 w0.35} & \textbf{MCUNet} & \textbf{ProxylessNAS w0.35}  \\
    
    \hline
    \hline

    ShiftAddNAS & Nan & Nan & Nan  \\

    Linear remap & Nan & Nan & Nan \\

    KL-loss only & 69.84 & 71.12   & 70.52  \\

    Linear + skip connect & \multirow{2}[1]{*}{71.02} & \multirow{2}[1]{*}{74.44} & \multirow{2}[1]{*}{72.99} \\
    + freeze &  & & \\

    \hline
    
    \textbf{Ours} & 71.83 & 74.59 & 73.86 \\
    \hline
    \hline
    \end{tabular}%
    }
  \label{tab:convergence}%
\end{table}%

%% file: tables/rebuttal_weightshare.tex
\begin{table}[htbp]
  \setlength{\tabcolsep}{4pt}
  \centering
  \caption{Ablation study on weight sharing. Augmented with Shift op. Tells why weight sharing is important.}
  \vspace{-1em}
  \renewcommand{\arraystretch}{1.25} 
  \resizebox{\linewidth}{!}{
    \begin{tabular}{l|c|c|c}
    \hline
    \hline
    \textbf{Method} & \textbf{MobileNetV2 w0.35}  & \textbf{MCUNet} & \textbf{ProxylessNAS w0.35}  \\
    
    \hline
    \hline

    Direct trained & 69.25	&70.87	&70.54	\\


    \textbf{w/o. weight sharing} & 
    \textbf{69.42}	&\textbf{71.09}	&\textbf{70.53}	 \\


    w/. weight sharing & 70.12	&72.68	&70.91	\\
        
    \hline
    \hline
    \end{tabular}%
    }
  \label{tab:rebuttal_weightsharing}%
  \vspace{0.0em}
\end{table}%

%% file: tables/rebuttal_woAug.tex
\begin{table}[h]
  \setlength{\tabcolsep}{4pt}
  \centering
  \caption{HWS is not a parameterization trick that can directly improve target model}
  \vspace{-1em}
  \renewcommand{\arraystretch}{1.25} 
  \resizebox{\linewidth}{!}{
    \begin{tabular}{c|c|c|c}
    \hline
    \hline
    \textbf{Method} & \textbf{MobileNetV2 w0.35} & \textbf{MCUNet} & \textbf{ProxylessNAS w0.35}  \\
    \hline
    \hline

    w/o. aug. , w/o. HWS & 69.25 & 70.87 & 70.54  \\
    \textbf{w/o. aug. , with HWS} & \textbf{68.32} & \textbf{71.13} & \textbf{69.88} \\
    with aug. , with HWS & 71.83 & 74.59 & 73.86\\
    \hline
    \hline
    \end{tabular}%
    }
  \label{tab:rebuttal_woAug}%
  \vspace{0.0em}
\end{table}%

%% file: tables/ablation_nas.tex
\begin{table}[h]
  \setlength{\tabcolsep}{4pt}
  \centering
  \caption{The ablation study of depth augmentation and block mutation in terms of accuracy on CIFAR100. Results are obtained under similar hardware cost.}
  \vspace{-1em}
  \renewcommand{\arraystretch}{1.25} 
  \resizebox{\linewidth}{!}{
    \begin{tabular}{c|ccc}
    \hline
    \hline
    \textbf{Method} & \textbf{Accuracy(\%)} & \textbf{Energy (mj)} & \textbf{Latency (ms)}  \\
    \hline
    \hline

    Aug. Width \& Expand  & 75.13 & 1.52 & 0.57 \\
    Aug. Width \& Expand \& Depth  & 75.63 & 1.42 & 0.66 \\
    Aug. Width \& Expand,  Mutation & 75.92 & 1.4 & 0.67 \\

    \hline

    \textbf{All} & 76.35 & 1.632 & 0.554 \\
        
    \hline
    \hline
    \end{tabular}%
  }
  \label{tab:ablation_nas}%
\end{table}%

%% file: tables/moreClassifyDatasets_moreResults.tex
\begin{table}[htbp]
  \vspace{0em}
  \setlength{\tabcolsep}{4pt}
  \centering
  \caption{Accuracy of MobileNetV3-w0.35 / ProxylessNAS-w0.35 / MobileNetV2-Tiny on more datasets.}
  \renewcommand{\arraystretch}{1.3} 
  \resizebox{\linewidth}{!}{
    \begin{tabular}{c|c|c|c|c|c}
    \hline
    \hline
    \multirow{2}[1]{*}{\textbf{Model}} & \multirow{2}[1]{*}{\textbf{Methods}} & \multirow{2}[1]{*}{\textbf{\space CIFAR10 \space }} & \multirow{2}[1]{*}{\textbf{ \space ImageNet \space }} & \multirow{2}[1]{*}{\textbf{ \space Food101 \space   }} & \multirow{2}[1]{*}{\textbf{\space Flower102 \space }} \\
      & & & & & \\
    \hline
    \hline 
    \textbf{MobileNetV3}
        & Shift               & 88.85  & 54.19 & 73.3 & 93.82   \\
    \textbf{w0.35}    & \textbf{AugShift}   & 92.83  & 56.07 & 75.1 & 96.28    \\
    \hline 
    \textbf{ProxylessNAS}
        & Shift              & 87.71  & 55.4 & 75.66 & 94.31    \\
    \textbf{w0.35}    & \textbf{AugShift}  & 92.37  & 56.3 & 77.02 & 96.76    \\
    \hline 
    \textbf{MobileNetV2}
        & Shift              & 88.02  & 48.92 & 72.91 & 94.63    \\
    \textbf{Tiny}    & \textbf{AugShift}  & 91.93  & 50.0 & 74.58 & 96.56    \\
    \hline
    \hline
    \end{tabular}%
  }
  \label{tab:MoreDatasets_moreResults}%
\end{table}%

%% file: sec/5_conclusion.tex
\section{Conclusion}

\indent 
\indent In this paper, ShiftAddAug is proposed for training multiplication-free tiny neural networks, which can improve accuracy without expanding the model size. It's achieved by putting the target multiplication-free tiny NN into a larger multiplicative NN to get auxiliary supervision. To relocate important weights into the target model, a novel heterogeneous weight sharing strategy is used to approach the weight discrepancy caused by inconsistent weight distribution. Based on the work above, a two stage neural architecture search is utilized to design more powerful models. Extensive experiments on image classification and semantic segmentation task consistently demonstrate the effectiveness of ShiftAddAug.